\pdfoutput=1
\documentclass[10pt,twocolumn,letterpaper]{article}

\usepackage{cvpr} 
\usepackage{times}
\usepackage{epsfig}
\usepackage{graphicx}
\usepackage{amsmath}
\usepackage{amssymb}
\usepackage[accsupp]{axessibility} 
\usepackage{cvpr}
\usepackage{times}
\usepackage{epsfig}
\usepackage{graphicx}
\usepackage{amsmath}
\usepackage{amssymb}
\usepackage{graphicx}
\usepackage{amsmath}
\usepackage{amssymb}
\usepackage{booktabs}
\usepackage{tabu}
\usepackage[numbers, sort]{natbib}

\usepackage[pagebackref]{hyperref}
\usepackage{xcolor}
\usepackage{array}
\usepackage{epsfig}
\usepackage{epstopdf}
\usepackage{booktabs}
\usepackage{algorithm}
\usepackage{algpseudocode}
\usepackage{lipsum}
\usepackage{multirow}
\usepackage{textcomp}
\usepackage{gensymb}
\usepackage{pifont}

\begin{document}

\title{CVVNet: A Cross-Vertical-View Network for Gait Recognition}

\author{
Xiangru Li\textsuperscript{1}\thanks{These authors contributed equally to this work.},
Wei Song\textsuperscript{1}\footnotemark[1]   \thanks{Corresponding author.},
Yingda Huang\textsuperscript{1},
Wei Meng\textsuperscript{1},
Le Chang\textsuperscript{1}\thanks{Corresponding author (Primary).},
Hongyang Li\textsuperscript{2}\\
\textsuperscript{1}Guangdong University of Technology, Guangzhou, China ,\\ \textsuperscript{2}Shanghai Innovation Institute, Shanghai, China, \\
\textit{\{2112404378, sonwe, 2112404370\}@mail2.gdut.edu.cn}, \\ \textit{meng0025@gdut.edu.cn}, \textit{lechang@gdut.edu.cn}, \textit{hongyang@sii.edu.cn}
}

\maketitle
\thispagestyle{empty}

\begin{abstract}
Gait recognition enables contact-free, long-range person identification that is robust to clothing variations and non-cooperative scenarios. While existing methods perform well in controlled environments, they struggle with cross-vertical view scenarios where elevation changes induce severe perspective distortions and self-occlusions. Our analysis reveals conventional CNN-based approaches relying on single-scale receptive fields fail to capture hierarchical frequency features, while standard self-attention mechanisms tend to over-smooth critical high-frequency edge patterns. To tackle this challenge, we propose Cross-Vertical-View Network(CVVNet), a frequency aggregation architecture specifically designed
for robust cross-vertical-view gait recognition. CVVNet employs a Multi-Scale Attention Gated Aggregation (MSAGA) module which consists of a High-Low Frequency Extraction module (HLFE) and a Dynamic Gated Aggregation (DGA) mechanism. Specifically, HLFE adopts parallel multi-scale convolution/max-pooling path and self-attention path as high- and low-frequency mixers for effective multi-frequency feature extraction from input silhouettes, while DGA is introduced to adaptively adjust the fusion ratio of frequency and spatial features. The integration of HLFE and DGA enables CVVNet to effectively handle distortions from view changes, significantly improving the recognition robustness across different vertical views. Experimental results show that our CVVNet achieves state-of-the-art performance on both datasets, delivering an 8.6\% improvement on DroneGait while retaining the leading position on Gait3D.


\end{abstract}

\section{Introduction}

Gait recognition is gaining attention for its non-contact, long-range biometric identification capabilities. As a unique motion pattern, it resists disguise, enables long-range acquisition, and is robust to clothing changes. These advantages make it highly promising for public security applications like border patrol, criminal investigation, and suspect tracking~\cite{sarkar2005humanid,nixon1996automatic,shen2024comprehensive,nixon2010human,wu2016comprehensive} .

\begin{figure}[t]
    \centering
    \includegraphics[width=\linewidth]{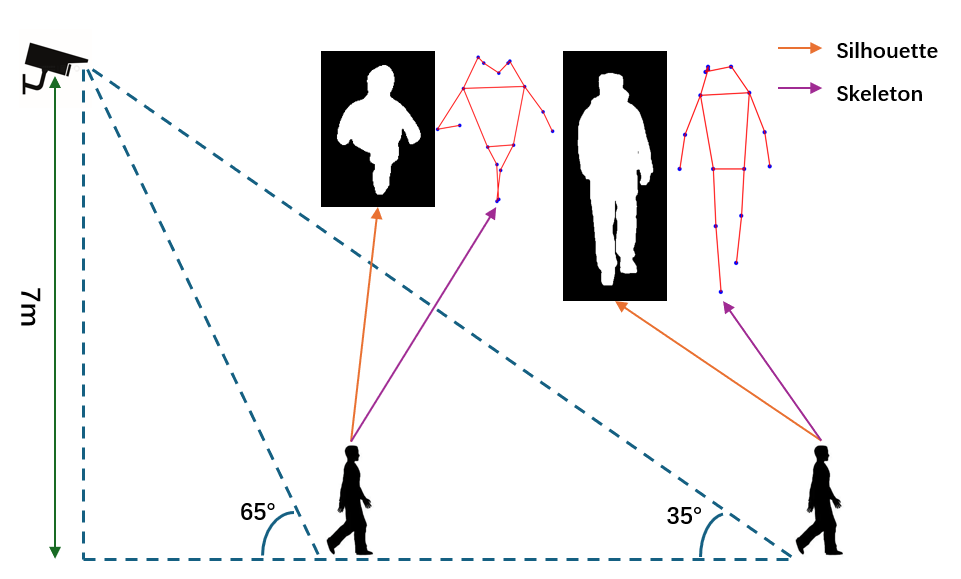}
    \caption{Cross-Vertical-View Gait Recognition: The Challenge}
    \label{fig:example}
\end{figure}

Deep learning-based approaches have dominated the state-of-the-art performance in widely adopted indoor gait recognition datasets such as CASIA-B~\cite{tan2006casia-b} and OU-MVLP~\cite{takemura2018ou-mvlp} by learning discriminative features directly from silhouette sequences~\cite{chao2019gaitset,fan2020gaitpart,zhang2021koopman,hou2020GLN,lin2021gaitgl}. 
While networks~\cite{chao2019gaitset,fan2020gaitpart,hou2020GLN,lin2021gaitgl,teepe2021gaitgraph,zhang2021koopman,zhang2023GaitTR} trained on these benchmarks demonstrate robust recognition capabilities within controlled indoor environments, they primarily focus on gait recognition under minimal vertical angle variations, where subjects and cameras remain at similar elevation levels. Consequently, existing models exhibit a tendency to overfit to the specific perspective heights represented in current datasets ~\cite{ma2023danet,kovacs2022vpnet}, limiting their generalizability to more diverse real-world scenarios.

In real-world surveillance scenarios, cameras are generally mounted at elevated positions~\cite{li2023dronegait}, creating complex viewing conditions. As illustrated in Fig.~\ref{fig:example}, during normal walking sequences, subjects experience continuous distance variations (far-to-near) and vertical angle change (low-to-high) with respect to the camera. This dynamic perspective induces significant deformation in critical anatomical components (e.g., foot trajectories and joint angles), while key body regions (e.g., torso and feet) suffer from severe self-occlusion under increased vertical angles.  Our experimental results demonstrate that under these challenging conditions, both skeleton-based and local texture-based feature extraction approaches exhibit substantially degraded performance. This perspective sensitivity constitutes a fundamental limitation that severely constrains the practical applicability of existing algorithms, presenting significant challenges to recognition robustness in pervasive cross-vertical viewing-angle scenarios in real-world applications.

Existing methods~\cite{chao2019gaitset,fan2020gaitpart,hou2020GLN,lin2021gaitgl} encounter two fundamental limitations under cross-vertical-view scenarios. First, convolutional networks with fixed receptive fields predominantly focus on single-scale details, resulting in insufficient capacity for modeling hierarchical frequency information. This shortage makes it challenging to maintain spatial consistency of contours under dynamic viewpoint changes, ultimately amplifying self-occlusion effects. Second, although self-attention mechanisms offer superior low-pass filtering characteristics that effectively capture low-frequency global feature, it suffers from an inherent constraint on preserving high-frequency motion details~\cite{park2022vision,si2022inception}. Such fragmented frequency representation creates a critical dilemma under continuous vertical viewpoint variations: convolutional architectures fail to compensate for cross-view deformations due to their weak sensitivity to global structural information, whereas self-attention mechanisms sacrifice fine-grained motion cues that are essential for discriminative identity features. The combined effect of these limitations substantially undermines the recognition robustness in complex outdoor surveillance environments, where vertical angle variations are prevalent and unavoidable.

To address the aforementioned challenges, we propose a Cross-Vertical-View Network (CVVNet) to address the generalization bottleneck of existing methods in cross-view scenarios by constructing dual mechanisms of multi-frequency feature separation and spatially adaptive gated fusion. The key component of CVVNet is the Multi-Scale Attention Gated Aggregation (MSAGA) module which consists of a High-Low-Frequency feature extraction (HLFE) module and a Dynamic Gated Aggregation (DGA) mechanism. In our HLFE module, the low-frequency branch employs spatial compression attention to disentangle the global structural components of feature contours, while the high-frequency branch utilizes depthwise separable convolution to extract local dynamic components. Furthermore, we use DGA to adaptively adjust the fusion ratio of frequency and spatial features.

Our main contributions are summarized as follows.
\begin{itemize}
\item We design a CVVNet to achieve high-low frequency feature representations for both cross-vertical-view and outdoor recognition scenarios.
\item We propose MSAGA to extract high-low frequency information from gait features and dynamically modulate the fusion ratios of different frequency components.
\item Extensive experiments demonstrate that our method achieves SOTA performance on both DroneGait~\cite{li2023dronegait} and Gait3D~\cite{zheng2022gait3d} datasets. Additionally, our rigorous ablation studies on DroneGait further validate the effectiveness of each component in CVVNet.
\end{itemize}

\begin{figure*}
\begin{center}
    \includegraphics[width=0.95\linewidth,height=2in]{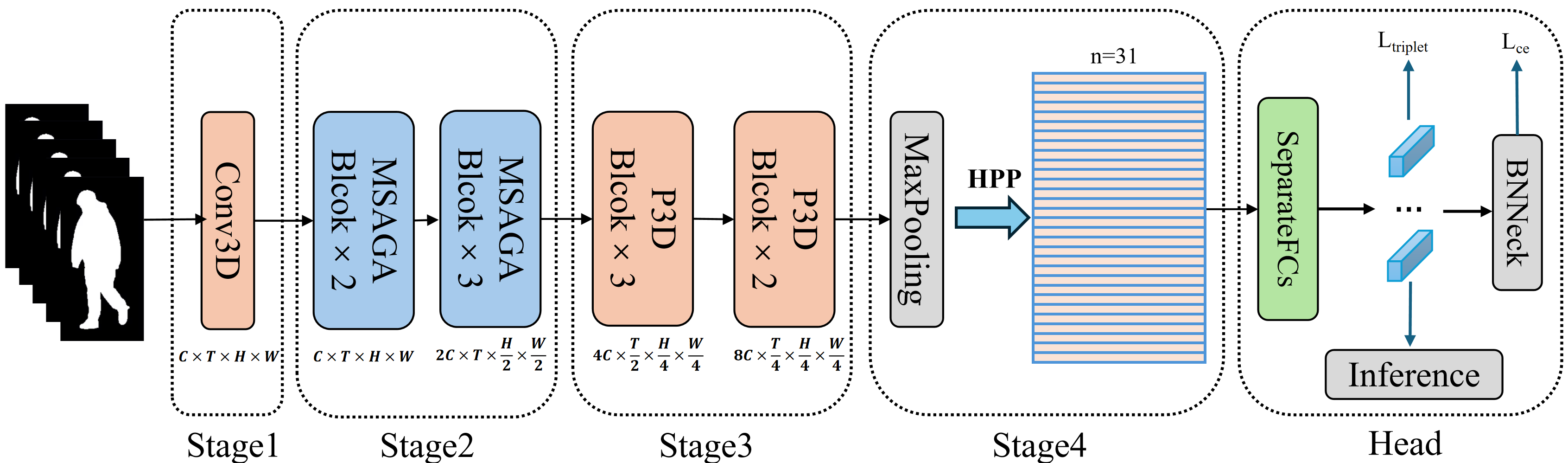}
\end{center}
   \caption{Overview of the whole CVVNet.}
\label{fig:pipeline}
\end{figure*}
\section{Related Work}

\subsection{Gait Recognition}
Gait recognition methods are classically divided into model-based and appearance-based approaches. 

Model-based methods employ structural representations (2D/3D skeletons, meshes) encoded through joint coordinates or vectors to reduce appearance dependency, theoretically improving robustness to covariates. Key advancements include PoseGait~\cite{liao2020posegait} (3D kinematics with engineered descriptors), GaitGraph~\cite{teepe2021gaitgraph} (graph networks for 2D poses), and HMRGait~\cite{li2020HMRGait} (SMPL-based transfer learning). Despite progress on controlled datasets like CASIA-B and OU-MVLP, model-based approaches remain inferior to appearance-based methods in real-world outdoor scenarios.

Appearance-based gait recognition methods use primarily informative visual characteristics to learn gait features from silhouette or RGB images. GaitSet~\cite{chao2019gaitset} innovatively treats the gait sequence as a set and employs a maximum function to compress the sequence of frame-level spatial features. GaitPart~\cite{fan2020gaitpart} suggested that gait models should focus on more local details and present the FConv layer. GaitGL~\cite{lin2021gaitgl} highlighted the importance of combining spatially local and global features for discriminative gait pattern extraction. Fan et al. ~\cite{fan2023opengait} thoroughly reviewed some of the above methods and further proposed a simple yet strong baseline model called GaitBase. DeepGaitV2~\cite{fan2023deepgaitv2} presented a unified perspective to explore how to construct deep models for state-of-the-art outdoor gait recognition. DANet~\cite{ma2023danet} created a dynamic attention mechanism between the features of neighboring pixels to learn more discriminative gait features. GaitGS~\cite{xiong2024gaitgs} captured multi-level motion patterns (micro and macro) and produces local-global temporal representations.

\subsection{Gait Transformers}

Vision transformer has achieved successful performance in many fields, such as semantic segmentation, objective detection and classification. Some works introduced transformer into the gait recognition field. Xu et al.~\cite{xu2020cross} developed a cross-view recognition model using pairwise spatial transformers, while Li et al.~\cite{li2019joint,li2021pose} proposed unified intensity transformers and cascade pose regression methods. Recent innovations include Hsu's GaitTAKE~\cite{hsu2022gaittake} integrating global-local temporal attention, Dou's MetaGait~\cite{dou2022metagait} addressing covariate conflicts through meta-knowledge injection, and Zou's multi-scale framework~\cite{zou2024multi} with attention-based feature fusion. In summary, the potential of deep gait transformers in addressing practical complexities becomes an attractive prospect.

Despite such emerging trend, existing gait recognition studies have not provided solid empirical evidence to fully validate the superiority of attention-based models in gait representation. This is mainly due to the inherent global attention mechanism of self-attention, which tends to prioritize low-frequency signals over high-frequency components during feature extraction~\cite{park2022vision,si2022inception}. However,  high-frequency information during continuous body movements in various gait input modalities (e.g., silhouettes), e.g., localized edge details and texture patterns, play a critical role in network performance. Thus, we propose an efficient feature extraction module that synergistically integrates the low-frequency modeling strengths of self-attention with the high-frequency capture capabilities of local convolutions. This framework encodes silhouette information as a composite of high- and low-frequency components, dynamically aggregates these complementary representations, and thus achieves context-aware calibration of multi-frequency features and enhancing the model's view generalization capability.

\begin{figure}
\begin{center}
    \includegraphics[width=0.8\linewidth,height=2.6in]{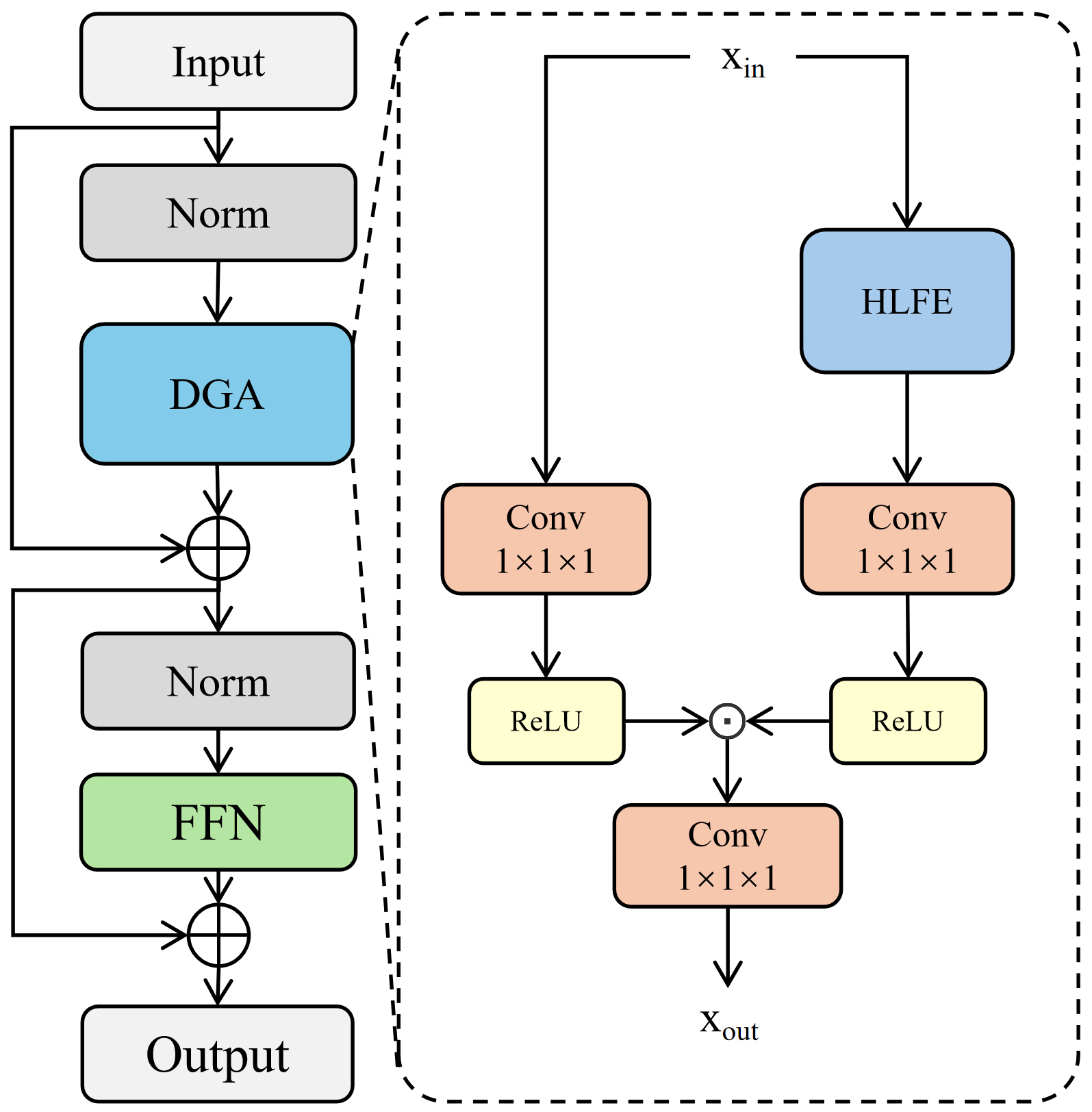}
\end{center}
   \caption{Multi-Scale Attention Gated Aggregation module (MSAGA) and Dynamic Gated Aggregation (DGA).}
\label{fig:MSAGA}
\end{figure}

\section{Method}\label{sec:method}

\begin{figure*}
\begin{center}
    \includegraphics[width=0.9\linewidth,height=3in]{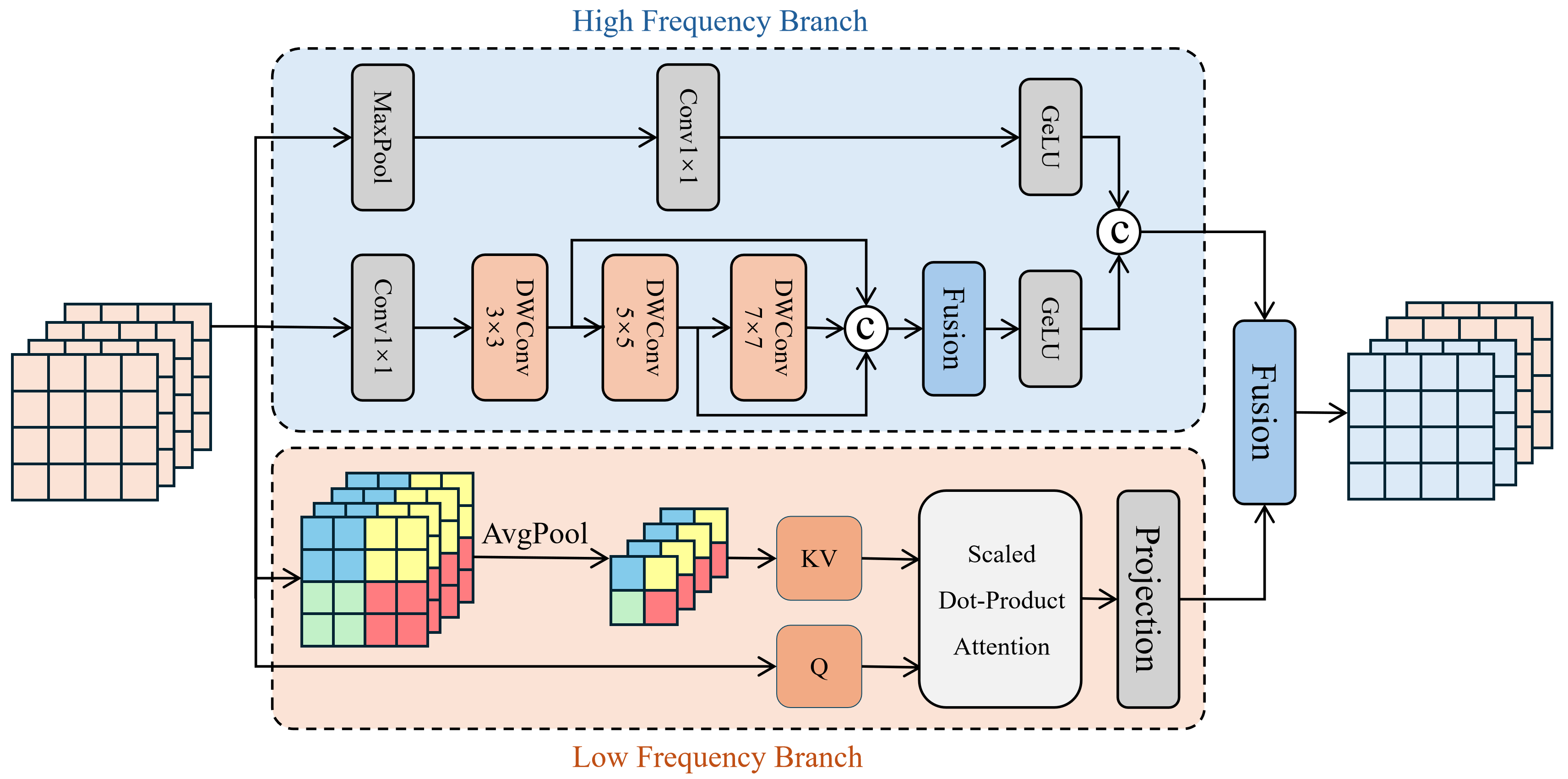}
\end{center}
   \caption{High-Low Frequency feature Extraction (HLFE) module.}
\label{fig:HLFE}
\end{figure*}

In this section, we first introduce the overview of our CVVNet framework. Then, we detail the High-Low Frequency feature Extraction module (HLFE) and Multi-Scale Attention Gated Aggregation module (MSAGA). Finally, we explain the optimization.

\subsection{Overview}\label{sec:pipeline}

The framework of the proposed method is illustrated in Fig.~\ref{fig:pipeline}. We first employ a 3D convolutional layer to extract shallow features. Subsequently, MSAGA is utilized to extract and aggregate both high- and low-frequency information from gait features. The processed features are then further refined through P3D blocks~\cite{fan2023deepgaitv2} to enhance their representations. Following this, a temporal max pooling operation is applied to aggregate the feature map sequence across the temporal dimension, outputting a holistic representation of the input gait sequence.

Next, the Horizontal Pyramid Pooling (HPP) module~\cite{fu2019HPP} hierarchically partitions the feature maps into multiple horizontal regions. Each region is aggregated into a feature vector which is subsequently mapped to a discriminative metric space using fully connected (FC) layers. Finally, the widely used BNNeck~\cite{luo2019BNNeck} is adopted to calibrate the feature space, and the training process is optimized via a joint loss function combining triplet loss and cross-entropy loss.

\subsection{Frequency Feature Extraction based on HLFE}\label{sec:HLFE}
As shown in Fig.~\ref{fig:HLFE}, HLFE consists of two branches for high frequency and low frequency feature extraction.

\textbf{High Frequency Branch.} We design a parallel structure to learn the high-frequency components in gait inputs, considering the high sensitivity of max pooling to salient features such as contours and textures, as well as the detail-perceiving ability of convolutional operations.

Pooling branch is the first branch to enhance spatially salient features through max pooling and projection:
\begin{align}
    \mathbf{X}_{\text{hf}} &= \text{MaxPool}(\mathbf{X})\\
    \mathbf{Y}_{\text{1}} &= \text{GELU}(\text{Conv}_{1\times1}(\mathbf{X}_{\text{hf}}))
\end{align}
where $\mathbf{X} \in \mathbb{R}^{B \times C \times H \times W}$ is the input feature map, $\mathbf{Y}_\text{1} \in \mathbb{R}^{B \times C \times H \times W}$.

The CNN branch is the second branch to employ cascaded multi-scale depthwise convolutions to capture hierarchical high-frequency details. Firstly, a $1\times1$ convolution is applied to transform the features and enhance their expressiveness:
\begin{equation}
    \mathbf{X}'_{\text{hf}} = \text{Conv}_{1\times1}(\mathbf{X})
\end{equation}
where $\mathbf{X}' \in \mathbb{R}^{B \times C \times H \times W}$. The transformed features are then processed through cascaded depthwise separable convolutions with progressively increasing kernel sizes:
\begin{align}
    \mathbf{F}_3 &= \text{DWConv}_{3\times3}(\mathbf{X}'_{\text{hf}}) \\
    \mathbf{F}_5 &= \text{DWConv}_{5\times5}(\mathbf{F}_3) \\
    \mathbf{F}_7 &= \text{DWConv}_{7\times7}(\mathbf{F}_5)
\end{align}
where $ \mathbf{F}_3, \mathbf{F}_5, \mathbf{F}_7\in \mathbb{R}^{B \times C \times H \times W} $, and $\text{DWConv}_{K\times K}$ denotes the depthwise convolution with kernel size $K$. This hierarchical design enables the network to capture high - frequency information across multiple scales. Then, the multi-scale features are concatenated and fused through a $1\times1$ convolution:
\begin{equation}
    \mathbf{F}_{\text{fused}} = \text{Conv}_{1\times1}\left(\text{Concat}(\mathbf{F}_3, \mathbf{F}_5, \mathbf{F}_7)\right)
\end{equation}
where $\mathbf{F}_{\text{fused}} \in \mathbb{R}^{B \times C \times H \times W}$. 
The Gaussian Error Linear Unit (GELU) activation introduces non-linearity while preserving edge continuity:
\begin{align}
    \mathbf{Y}_{\text{2}} &= \text{GELU}(\mathbf{F}_{\text{fused}}) \\
    \mathbf{Y}_{\text{high}} &= \text{Concat}(\mathbf{Y}_1, \mathbf{Y}_2)
\end{align}
 where $\mathbf{Y}_{\text{2}}\in \mathbb{R}^{B \times C \times H \times W}$, and $\mathbf{Y}_{\text{high}}\in \mathbb{R}^{B \times 2C \times H \times W}$ is the final output of high-frequency branch.

\textbf{Low Frequency Branch.} Given silhouette features $\mathbf{X} \in \mathbb{R}^{B \times C \times H \times W}$, we reshape it for attention computation:
    \begin{equation}
        \mathbf{X}' = \text{Permute}(\mathbf{X})
    \end{equation}
    where $\mathbf{X}'\in \mathbb{R}^{B \times H \times W \times C}$. $B$, $C$, $H$, and $W$ denote batch size, channels, height, and width, respectively.

We retain spatial details via linear projection to localize body parts that maintain relative positions across views:
    \begin{equation}
        \mathbf{Q} = \mathbf{W}_q \mathbf{X}'
    \end{equation}
    where $\mathbf{Q}\in \mathbb{R}^{B \times H \times W \times C}$. $\mathbf{W}_q \in \mathbb{R}^{C \times C}$ is the learnable query projection matrix. The query projection preserves the original spatial resolution.

Average Pooling is adopted to extract low-frequency features while preserving torso inclination and gait cycle phases. Projections $\mathbf{W}_k, \mathbf{W}_v \in \mathbb{R}^{C \times C}$ encode viewpoint-robust structural semantics.:
    \begin{align}
        \mathbf{X}_{\text{down}} &= \text{AvgPool}(\mathbf{X}'; s=2)  \\
        \mathbf{K}, \mathbf{V} &= \mathbf{W}_k \mathbf{X}_{\text{down}}, \mathbf{W}_v \mathbf{X}_{\text{down}} 
    \end{align}
    where $\mathbf{X}_{\text{down}}\in \mathbb{R}^{B \times C \times H/2 \times W/2}$, $\mathbf{K} \in \mathbb{R}^{B \times (H/2\times W/2) \times C}$, $\mathbf{V} \in \mathbb{R}^{B \times (H/2\times W/2) \times C}$, $s=2$ is the downsampling stride, and $\mathbf{W}_k, \mathbf{W}_v \in \mathbb{R}^{C \times C}$ are key/value projection matrices. Then We calculate cross view correlations using the attention mechanism: \begin{equation}
        \text{Attention}(\mathbf{Q}, \mathbf{K}, \mathbf{V}) = \text{Softmax}\left(\frac{\mathbf{Q}\mathbf{K}^\top}{\sqrt{d}}\right)\mathbf{V}
    \end{equation}
where $d = C/N_h$ is the scaled dimension with $N_h=8$ attention heads. Here, $N_h$ attention heads model multi-scale spatial relationships across viewpoints. The $\sqrt{d}$ scaling stabilizes gradient propagation. Then, fuse multi-head outputs:
    \begin{equation}
        \mathbf{Y}_{\text{low}} = \mathbf{W}_o \text{Concat}(\text{Head}_1, ..., \text{Head}_{N_h})
    \end{equation}
    where $\mathbf{Y}_{\text{low}} \in \mathbb{R}^{B \times C \times H \times W}$, $\mathbf{W}_o \in \mathbb{R}^{C \times C}$ is the output projection matrix.

Finally, the high and low frequency information extracted are concatenated along the channel dimension and fused via a $1\times1$ convolution, with identity information preserved:
    \begin{equation}
    \mathbf{Y}_{\text{out}} = \text{Conv}_{1\times1}\left(\text{Concat}(\mathbf{Y}_{high}, \mathbf{Y}_{low})\right)+\mathbf{X}
    \end{equation}
where $\mathbf{Y}_\text{out}\in \mathbb{R}^{B \times C \times H \times W}$ is the final output of HLFE.

\subsection{Dynamic Feature Aggregation with MSAGA}\label{sec:MSAGA}

Then, the MSAGA dynamically fuses hierarchical features through the Dynamic Gated Aggregation (DGA) mechanism. As shown in Fig.~\ref{fig:MSAGA}, given input features $\mathbf{X} \in \mathbb{R}^{B \times C \times T \times H \times W}$ where $B,C,T,H,W$ denote batch size, channels, temporal length, height, and width, the module operates as:
\begin{align}
\mathbf{G} & = \mathbf{W}_g  \mathbf{X} \\
\mathbf{V}_{\text{HL}} &= \mathcal{F}_{\text{HL}}(\mathbf{X}) \\
\mathbf{V} &= \mathbf{W}_v \mathbf{V}_{\text{HL}}
\end{align}
where $\mathbf{W}_g, \mathbf{W}_v \in \mathbb{R}^{C \times C \times 1 \times 1 \times 1}$ are 3D convolution weights for gating and value projection respectively, and $\mathcal{F}_{\text{HL}}$ denotes the HLFE. 

The final output is computed through activation-controlled gating:
\begin{equation}
\mathbf{Y} = \mathbf{W}_p \left(\text{ReLU}(\mathbf{G}) \odot \text{ReLU}(\mathbf{V})\right)
\end{equation}
where $\odot$ denotes element-wise multiplication, and $\mathbf{W}_p \in \mathbb{R}^{C \times C \times 1 \times 1 \times 1}$ is the projection convolution.

\subsection{Optimization}
In the training stage of CVVNet, a combined loss function consisting of triplet loss ($\mathcal{L}_{tri}$) and cross-entropy loss ($\mathcal{L}_{ce}$) is calculated to supervise the learning process:
\begin{equation}
    \mathcal{L}=\alpha\mathcal{L}_{tri}+\beta\mathcal{L}_{ce},
    \label{equ:total loss}
\end{equation}
where $\alpha$ and $\beta$ are hyper-parameters to balance the contributions to the total loss $\mathcal{L}$.
\section{Experiments and Results}

In this section, we first introduce the datasets and implementation details. Then, we conduct comparisons between our proposed CVVNet and state-of-the-art gait recognition methods under both same-vertical-view and cross-vertical-view scenarios, with detailed analysis of the results. Furthermore, we conduct extensive ablation studies that conclusively demonstrate the effectiveness of each component in CVVNet.

\begin{table*}[htbp]
    \setlength{\abovecaptionskip}{2pt}
    \setlength{\belowcaptionskip}{2pt}
    \fontsize{5}{6}\selectfont
    \centering
    \vspace{1mm}
    \renewcommand{\arraystretch}{1.0}
    \resizebox{16cm}{!}{%
    \begin{tabular}{c|c|ccccccccc}
        \toprule
        \multirow{4}{*}{Train} & \multirow{4}{*}{Method} & \multicolumn{9}{c}{Test View (\%)} \\
        \cline{3-11} & & \multicolumn{3}{c}{\multirow{2}{*}{Low}} & \multicolumn{3}{c}{\multirow{2}{*}{Mid}} & \multicolumn{3}{c}{\multirow{2}{*}{High}} \\
        & & \multicolumn{3}{c}{} & \multicolumn{3}{c}{} & \multicolumn{3}{c}{} \\
        \cline{3-11} & & NM & BG & CL & NM & BG & CL & NM & BG & CL \\
        \hline
        \multirow{10}{*}{Low} 
        & GaitSet \cite{chao2019gaitset} 
        & 99.5 & 99.5 & 99.8 & 72.5 & 74.7 & 71.6 & 20.5 & 23.7 & 22.5 \\ 
        & GaitPart \cite{fan2020gaitpart} 
        & 99.2 & 99.3 & 99.7 & 63.4 & 64.8 & 63.8 & 17.8 & 20.2 & 20.2 \\
        & GaitGL \cite{lin2021gaitgl} 
        & 99.5 & 99.6 & 99.7 & 84.6 & 84.7 & 81.7 & 24.8 & 26.3 & 27.2 \\
        & GaitTR \cite{zhang2023GaitTR}
        & 84.9 & 81.7 & 81.1 & 6.3 & 5.6 & 4.9 & 7.7 & 7.9 & 8.1 \\
        & GaitBase \cite{fan2023opengait} 
        & 99.6 & 99.5 & 99.0 & 86.9 & 88.6 & \textbf{85.5} & 20.6 & 23.6 & 23.4 \\
        & DeepGaitv2 \cite{fan2023deepgaitv2} 
        & 99.8 & \textbf{99.9} & 99.3 & 88.8 & 84.8 & 82.6 & 25.7 & 30.2 & 28.7 \\
        & GaitGS \cite{xiong2024gaitgs} 
        & 99.5 & 99.7 & 99.6 & 85.8 & 87.3 & 83.1 & 21.8 & 23.3 & 24.2 \\
        & QAGait \cite{wang2024qagait} 
        & \textbf{99.9} & 99.8 & 99.4 & 87.3 & 89.0 & 85.1 & 22.1 & 24.5 & 23.1 \\
        & GLGait \cite{peng2024glgait}
        & 97.6 & 98.1 & 98.9 & 82.2 & 82.4 & 79.1 & 21.7 & 25.2 & 24.1 \\ 
        & Ours 
        & 99.6 & 99.7 & \textbf{99.9} & \textbf{88.9} & \textbf{91.2} & \textbf{85.5} & \textbf{34.3} & \textbf{36.0} & \textbf{36.7} \\ 
        \hline
        \multirow{10}{*}{Mid}
        & GaitSet \cite{chao2019gaitset} 
        & 94.6 & 93.5 & 96.9 & 94.3 & 97.2 & 92.7 & 37.6 & 39.1 & 38.3 \\ 
        & GaitPart \cite{fan2020gaitpart} 
        & 95.2 & 94.9 & 97.3 & 95.7 & 98.7 & 93.9 & 32.8 & 32.8 & 33.9 \\
        & GaitGL \cite{lin2021gaitgl}
        & 95.3 & 96.1 & 97.2 & 95.6 & 98.1 & 92.1 & 46.1 & 48.5 & 47.4 \\
        & GaitTR \cite{zhang2023GaitTR}
        & 16.1 & 15.7 & 14.1 & 60.1 & 58.8 & 57.9 & 14.9 & 17.9 & 15.9 \\
        & GaitBase \cite{fan2023opengait} 
        & 98.7 & 99.1 & 99.2 & 96.8 & \textbf{99.8} & 95.0 & 42.9 & 44.2 & 46.4 \\
        & DeepGaitv2 \cite{fan2023deepgaitv2} 
        & \textbf{98.9} & 99.3 & 99.3 & 96.4 & 99.6 & 94.9 & 51.5 & 51.9 & 56.3\\
        & GaitGS \cite{xiong2024gaitgs} 
        & 95.7 & 95.9 & 97.2 & 95.5 & 97.4 & 93.7 & 41.8 & 41.9 & 42.2 \\
        & QAGait \cite{wang2024qagait} 
        & 98.7 & 98.9 & 98.4 & 96.5 & \textbf{99.8} & \textbf{95.8} & 45.0 & 46.6 & 45.9 \\
        & GLGait \cite{peng2024glgait}
        & 96.5 & 96.1 & 96.9 & 94.8 & 95.8 & 93.7 & 55.0 & 53.1 & 57.6 \\ 
        & Ours 
        & 98.5 & \textbf{99.5} & \textbf{99.5} & \textbf{97.3} & 99.7 & 95.7 & \textbf{55.3} & \textbf{55.5} & \textbf{58.6} \\ 
        \hline
        \multirow{10}{*}{High}
        & GaitSet \cite{chao2019gaitset} 
        & 43.8 & 46.4 & 51.1 & 40.4 & 44.8 & 47.7 & 58.3 & 55.1 & 60.9 \\ 
        & GaitPart \cite{fan2020gaitpart} 
        & 48.8 & 51.6 & 55.9 & 41.7 & 43.8 & 47.4 & 68.1 & 79.1 & 75.8 \\
        & GaitGL \cite{lin2021gaitgl}
        & 52.4 & 54.9 & 59.1 & 49.1 & 48.3 & 50.9 & 54.9 & 55.2 & 56.7 \\
        & GaitTR \cite{zhang2023GaitTR}
        & 7.1 & 8.1 & 7.0 & 11.5 & 14.7 & 12.6 & 46.7 & 52.2 & 50.3 \\
        & GaitBase \cite{fan2023opengait} 
        & 71.4 & 73.1 & 77.5 & 66.8 & 70.0 & 72.2 & 83.2 & 88.2 & 84.3 \\
        & DeepGaitv2 \cite{fan2023deepgaitv2} 
        & 79.8 & 82.1 & 85.8 & 73.5 & 76.0 & 80.0 & 85.2 & \textbf{88.8} & 86.4 \\
        & GaitGS \cite{xiong2024gaitgs} 
        & 49.5 & 51.6 & 55.5 & 45.1 & 47.4 & 51.4 & 69.5 & 73.7 & 71.1 \\
        & QAGait \cite{wang2024qagait} 
        & 67.2 & 67.0 & 74.2 & 63.8 & 67.7 & 69.3 & 85.1 & 90.6 & 86.8 \\
        & GLGait \cite{peng2024glgait}
        & 69.8 & 73.8 & 78.9 & 72.9 & 74.1 & 77.5 & 78.1 & 77.7 & 78.7 \\ 
        & Ours 
        & \textbf{81.1} & \textbf{84.4} & \textbf{87.7} & \textbf{76.7} & \textbf{78.9} & \textbf{81.3} & \textbf{86.4} & 88.5 & \textbf{87.3} \\ 
        \bottomrule 
    \end{tabular}
    }
    \vspace{4mm} 
    \caption{Cross-view gait recognition performance on DroneGait dataset (Rank-1 accuracy).}
    \label{tab:drone}
    \vspace{-0.3cm}
\end{table*}

\subsection{Datasets and Implementation Details}

The dataset information and implementation details in our experiments are as follows.

\textbf{DroneGait}~\cite{li2023dronegait} is a drone-based gait dataset focusing on high vertical view recognition, with 96 subjects and 22,000 sequences with vertical angles from $0^\circ$ to $80^\circ$. It categorizes data into three view groups: low ($0^\circ$), medium ($30^\circ\text{--}60^\circ$), and high ($60^\circ\text{--}80^\circ$), with frame-level alignment for cross-view analysis. Following the standard protocol, we split the data into 74 training and 26 testing subjects, evaluating recognition performance through probe-gallery matching where gallery sequences cover multi-view conditions.

\textbf{Gait3D}~\cite{zheng2022gait3d} is a large-scale comprehensive dataset for gait recognition, containing silhouettes, 2D/3D human poses, and 3D meshes. It comprises 4,000 subjects, 25,309 sequences, and 39 distinct camera viewpoints. Collected in supermarket environments with significant high vertical views, the dataset contains abundant human silhouettes from elevated perspectives, presenting enhanced challenges for gait recognition tasks. Following the official protocol, we use 3,000 subjects for training and 1,000 for testing, evaluating performance via rank-1/5 accuracy and mAP metrics computed from query-gallery sequence similarities.

\textbf{Implementation Details.} Our network implementation and reproduction experiments of state-of-the-art models are both deployed through the OpenGait codebase~\cite{fan2023opengait}, and are conducted on a system with four NVIDIA RTX 3090 GPUs. We first integrate the DroneGait dataset into this framework, following the preprocessing methodology described in~\cite{chao2019gaitset}. The input silhouettes are aligned and resized to $64 \times 44$ pixels, and then organized into sequences of length $30$ as model inputs. Both the loss functions and data augmentation strategies of CVVNet adhere to the configurations established in GaitBase~\cite{fan2023opengait}. To ensure fair training, we set the batch size for all models to 8×16 for DroneGait and 32×4 for Gait3D, while keeping other settings consistent with the original reproduced models.

The optimization setup uses AdamW with base learning rate $1\times10^{-4}$ and weight decay $0.05$, coupled with OneCycleLR scheduler configured with maximum learning rate $6\times10^{-4}$, total training steps $80,\!000$, and warm-up phase covering $6\%$ of the cycle. In particular, identical optimizer and scheduler configurations are strictly maintained for both Gait3D and DroneGait datasets to ensure methodological consistency.


\begin{table*}[htbp]
    \setlength{\abovecaptionskip}{2pt}
    \setlength{\belowcaptionskip}{2pt}
    \fontsize{5}{6}\selectfont
    \renewcommand{\arraystretch}{1.0}
    \centering
    \vspace{1mm}
    \newlength{\tablewidth}
    \newlength{\tableheight}
    \setlength{\tablewidth}{0.8\textwidth}
    \setlength{\tableheight}{2.4cm}
    \resizebox{\tablewidth}{\tableheight}{%
    \begin{tabular}{c|ccccccccc}
        \toprule
        \multirow{3}{*}{Method} & \multicolumn{9}{c}{Test View (\%)} \\
        \cline{2-10} & \multicolumn{3}{c}{\multirow{2}{*}{Low}} & \multicolumn{3}{c}{\multirow{2}{*}{Mid}} & \multicolumn{3}{c}{\multirow{2}{*}{High}} \\
        & \multicolumn{3}{c}{} & \multicolumn{3}{c}{} & \multicolumn{3}{c}{} \\
        \cline{2-10} & NM & BG & CL & NM & BG & CL & NM & BG & CL \\
        \hline
        GaitSet \cite{chao2019gaitset} 
        & 94.8 & 95.6 & 98.2 & 85.6 & 87.1 & 87.0 & 32.1 & 33.8 & 30.9 \\ 
        GaitPart \cite{fan2020gaitpart} 
        & 92.8 & 93.3 & 94.1 & 80.4 & 80.0 & 82.1 & 27.3 & 30.5 & 29.2 \\
        GaitGL \cite{lin2021gaitgl} 
        & 93.6 & 94.8 & 94.6 & 83.8 & 84.1 & 81.9 & 32.6 & 36.2 & 34.8 \\
        GaitBase \cite{fan2023opengait} 
        & 97.9 & 98.9 & 98.1 & 95.3 & 98.0 & 93.7 & 42.2 & 43.4 & 44.0 \\
        DeepGaitv2 \cite{fan2023deepgaitv2} 
        & 95.2 & 97.6 & 97.3 & \textbf{95.9} & \textbf{98.4} & 94.9 & 45.2 & 47.6 & 47.3 \\
        QAGait \cite{wang2024qagait} 
        & 98.0 & 98.6 & 97.9 & 95.7 & 97.8 & 93.3 & 38.3 & 39.2 & 38.9 \\ \midrule
        Ours 
        & \textbf{98.4} & \textbf{99.5} & \textbf{98.8} & 95.8 & 97.4 & \textbf{95.6} & \textbf{45.9} & \textbf{48.4} & \textbf{47.5} \\ 
        \bottomrule 
    \end{tabular}
    }
    \vspace{2mm} 
    \caption{Train on Gait3D, test on DroneGait-High}
    \vspace{-5mm}
    \label{tab:gait3d-drone}
\end{table*}

\subsection{Results and Analysis}\label{sec:res}
To verify the effectiveness of our method, we introduce several latest gait recognition methods for comparison, including GaitSet~\cite{chao2019gaitset}, GaitPart~\cite{fan2020gaitpart}, GaitGL~\cite{lin2021gaitgl}, GaitBase~\cite{fan2023opengait}, DeepGaitV2~\cite{fan2023deepgaitv2}, GaitGS~\cite{xiong2024gaitgs}, QAGait~\cite{wang2024qagait}, and GLGait~\cite{peng2024glgait}.

\textbf{Evaluation on DroneGait.}
We extensively evaluate CVVNet's view generalization capability on the DroneGait dataset under both cross-view and matched-view scenarios. As detailed in Tab~\ref{tab:drone}, CVVNet demonstrates remarkable robustness against extreme perspective shifts. In the challenging low-to-high cross-view scenario, our method achieves accuracy of 34.3\% NM, 36.0\% BG and 36.7\% CL, surpassing DeepGaitv2 by significant margins of 8.6\%, 5.8\%, and 8.0\%, respectively. This breakthrough stems from the dual-branch architecture's ability to suppress view-specific distortions through low-frequency global modeling while enhancing discriminative local patterns via high-frequency refinement. For mid- to high-view adaptation, CVVNet achieves balanced improvements of 3.8\% to 4.0\% over GLGait in all conditions, reaching 55.3\% NM, 55.5\% BG and 58.6\% CL accuracy. 

Under matched-view conditions, CVVNet maintains state-of-the-art performance, achieving a near-perfect accuracy of 99.6\% NM, 99.7\% BG and 99.9\% CL at low vertical views. For mid-view recognition, the method reaches 97.3\% NM accuracy, surpassing GaitBase by 0.5\%, while marginally trailing the SOTA methods by 0.1\% and 0.1\% in the BG and CL metrics, respectively. In high vertical view scenarios, our method leads SOTA methods by 1.2\% in NM and 0.5\% in CL, and trails by only 0.3\% in BG. These experimental results collectively validate CVVNet's recognition robustness and consistent superiority in both ideal and real-world surveillance scenarios.

\textbf{Evaluation on Gait3D.}
The results on the Gait3D dataset further validate our hypothesis: enhancing viewpoint generalization directly translates to improved accuracy in complex outdoor environments. As shown in Tab~\ref{table:gait3d}, CVVNet achieves state-of-the-art performance across all metrics, attaining 75.6\% Rank-1, 89.1\% Rank-5, and 68.3\% mAP. This represents a 1.2\% Rank-1 improvement over DeepGaitV2 (74.4\%) and a 2.5\% mAP gain, demonstrating particularly strong advantages in handling real-world variations like occlusions and viewpoint shifts.

\begin{table}
	\centering
        \footnotesize
	\begin{tabular}{c|c|c|c|c}
		\toprule
		\cmidrule(r){1-5}
		Method     & Publication      & R-1    & R-5   & mAP \\
		\midrule
		GaitSet~\cite{chao2019gaitset}  & AAAI 2019   & 36.7 & 58.3 & 30.0 \\
        GaitPart~\cite{fan2020gaitpart} & CVPR 2020   & 28.2 & 47.6 & 21.6 \\
        GLN     ~\cite{hou2020GLN}      & ECCV 2020   & 31.4 & 52.9 & 24.7 \\
        GaitGL  ~\cite{lin2021gaitgl}   & ICCV 2021   & 63.8 & 80.5 & 55.9 \\
        GaitGCI ~\cite{dou2023gaitgci}  & CVPR 2023   & 50.3 & 68.5 & 39.5 \\
        GaitBase~\cite{fan2023opengait} & CVPR 2023   & 64.6 & 73.9 & 45.5 \\
        GaitRef ~\cite{zhu2023gaitref}  & IJCB 2023   & 49.0 & 69.3 & 40.7 \\
  DeepGaitV2 ~\cite{fan2023deepgaitv2}  & Arxiv 2023  & 74.4 & 88.0 & 65.8 \\
        GaitGS ~\cite{xiong2024gaitgs}  & ICIP 2024   & 54.2 & {-}  &  {-} \\
        QAGait ~\cite{wang2024qagait}   & AAAI 2024   & 67.0 & 81.5 & 56.5 \\
        GLGait ~\cite{peng2024glgait}   & ACM 2024    & \textbf{77.6} & 88.4 & \textbf{69.6} \\ \midrule
        {Ours}   & Arxiv 2025  & 75.6 & \textbf{89.1} & 68.3 \\
		\bottomrule
	\end{tabular}
        \vspace{2mm} 
	\caption{Comparison of Rank-1, 5 and mean Average Precision(\%) on Gait3D dataset.}
         \vspace{-5mm}
\label{table:gait3d}
\end{table}

\begin{table}
	\centering
        \footnotesize
	\begin{tabular}{c|c|c|c|c}
		\toprule
		\cmidrule(r){1-5}
		Method     & Publication      & R-1    & R-5   & mAP \\
		\midrule
		GaitSet~\cite{chao2019gaitset}  & AAAI 2019   & 7.2 & 15.9 & 5.7 \\
        GaitPart~\cite{fan2020gaitpart} & CVPR 2020   & 8.4 & 18.2 & 6.3 \\
        GaitGL  ~\cite{lin2021gaitgl}   & ICCV 2021   & 18.2 & 30.7 & 11.9 \\
        GaitBase~\cite{fan2023opengait} & CVPR 2023   & 14.3 & 26.1 & 10.2 \\
  DeepGaitV2 ~\cite{fan2023deepgaitv2}  & Arxiv 2023  & 20.5 & 35.8 & 13.9 \\
        GaitGS ~\cite{xiong2024gaitgs}  & ICIP 2024   & - & {-} &  {-} \\
        QAGait ~\cite{wang2024qagait}   & AAAI 2024   & 15.5 & 27.8 & 8.9 \\
        GLGait ~\cite{peng2024glgait}   & ACM 2024    & 15.0 & 27.4 & 10.2 \\ \midrule
        {Ours}   & Arxiv 2025  & \textbf{20.9} & \textbf{36.0} & \textbf{14.2} \\
		\bottomrule
	\end{tabular}
        \vspace{2mm} 
	\caption{Train on DroneGait-Low, test on Gait3D.}
        \vspace{-5mm}
\label{table:drone-gait3d}
\end{table}

\begin{figure*}
\begin{center}
    \includegraphics[width=\linewidth,height=1in]{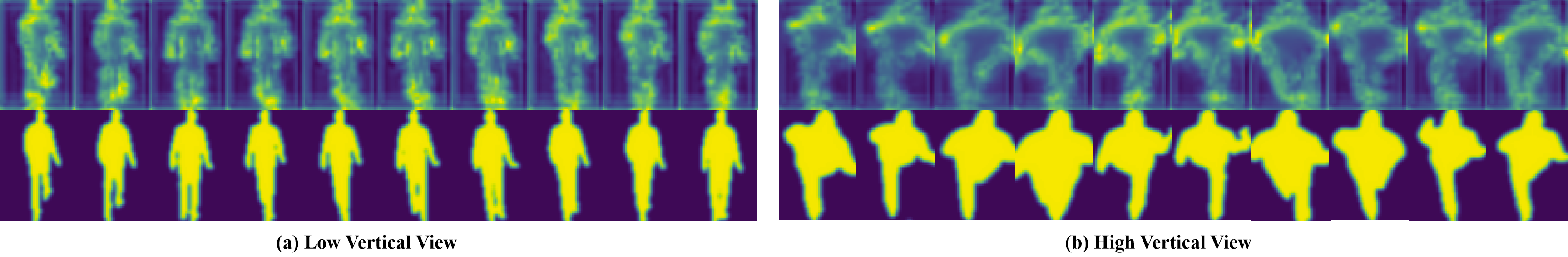}
\end{center}
   \caption{Heatmap Analysis of Feature Activation Patterns in CVVNet Under Different Vertical Views.}
   \vspace{-3mm}
\label{fig:heatmap}
\end{figure*}

\textbf{Cross-domain Evaluation.}
Our bidirectional cross-dataset evaluation between Gait3D and DroneGait validates CVVNet's superior domain generalization. 
As show in Tab~\ref{tab:gait3d-drone} and Tab~\ref{table:drone-gait3d}, when trained on DroneGait-Low and tested on Gait3D, CVVNet achieves a state-of-the-art 20.9\% Rank-1 accuracy, surpassing DeepGaitV2's previous best result (20.5\%) by 0.4\% while attaining 14.2\% mAP—a 0.3\% improvement over DeepGaitV2. Notably, it outperforms QAGait by 5.4\% absolute Rank-1 accuracy, confirming effective transfer of drone-learned viewpoint invariance to outdoor scenes. In the reverse evaluation trained on Gait3D and tested on DroneGait-High, CVVNet demonstrates consistent superiority: 45.9\% NM, 48.4\% BG, and 47.5\% CL accuracy under high-vertical-view conditions, exceeding DeepGaitv2 by 0.7-0.8\% across metrics. For low-vertical-view scenarios, it reaches 98.4\% NM and 99.5\% BG accuracy, outperforming GaitBase's prior records by 1.2\%. This bidirectional dominance stems from our dual-branch architecture's adaptive feature disentanglement, effectively bridging drone-ground domain gaps.

\begin{table}[htbp]
    \centering
    \label{tab:gait_recognition}
    \resizebox{0.4\textwidth}{!}{
    \fontsize{5}{6}\selectfont
    \begin{tabular}{c|c|c|c|c}
        \toprule
        FE. & Aggr. & NM & BG & CL \\
        \midrule
        P3D & Add & 30.3 & 32.2 & 33.6 \\
        P3D & Concat & 31.5 & 33.2 & 34.3 \\
        P3D & DGA & 33.4 & 34.5 & 35.5 \\ \midrule
        HLFE & Add & 31.4 & 32.3 & 33.7 \\ 
        HLFE & Concat & 33.0 & 34.7 & 35.5 \\ 
        HLFE & DGA & \textbf{34.3} & \textbf{36.0} & \textbf{36.7} \\
        \bottomrule
    \end{tabular}
    }
    \vspace{4mm} 
\caption{Ablation study of the proposed modules on the DroneGait dataset across Low to High cross-view scenarios, showing rank-1 accuracy (\%). FE. denotes the feature extraction module, and Aggr. represents the feature aggregator.}
\vspace{-3mm}
\label{table:ablation}
\end{table}

\subsection{Ablation studies}

To quantify the contributions of the HLFE and DGA components in MSAGA, we conduct ablation studies on the DroneGait dataset under low-to-high vertical view conditions. As shown in Tab~\ref{table:ablation}, HLFE consistently demonstrated superior performance compared to the P3D baseline. Across all aggregation strategies, configurations employing HLFE consistently outperformed their respective P3D counterparts. Notably, in the most challenging CL cross-view scenario, the HLFE-DGA model achieved a Rank-1 accuracy of 36.7\%, representing a 3.1\% improvement over P3D-DGA (33.6\%). This highlights the effectiveness of HLFE in extracting and encoding crucial low-frequency gait information, particularly under complex viewpoint variations.

The DGA mechanism demonstrates critical advantages over static fusion methods. Replacing traditional fusion methods with DGA consistently resulted in accuracy improvements ranging from 1.3\% to 3.1\% across different feature extractors (P3D or HLFE) and various view scenarios. For instance, within the HLFE series, DGA improved accuracy by 1.3\% over Concat (from 33.0\% to 34.3\%) and by 2.9\% over Add (from 31.4\% to 34.3\%) in the NM scenario. This result strongly confirms DGA's ability to dynamically weight multi-frequency features based on context, rather than applying fixed fusion rules, thereby enabling more effective information integration.

\subsection{Visualization }

\begin{figure}
\begin{center}
    \includegraphics[width=\linewidth,height=1.5in]{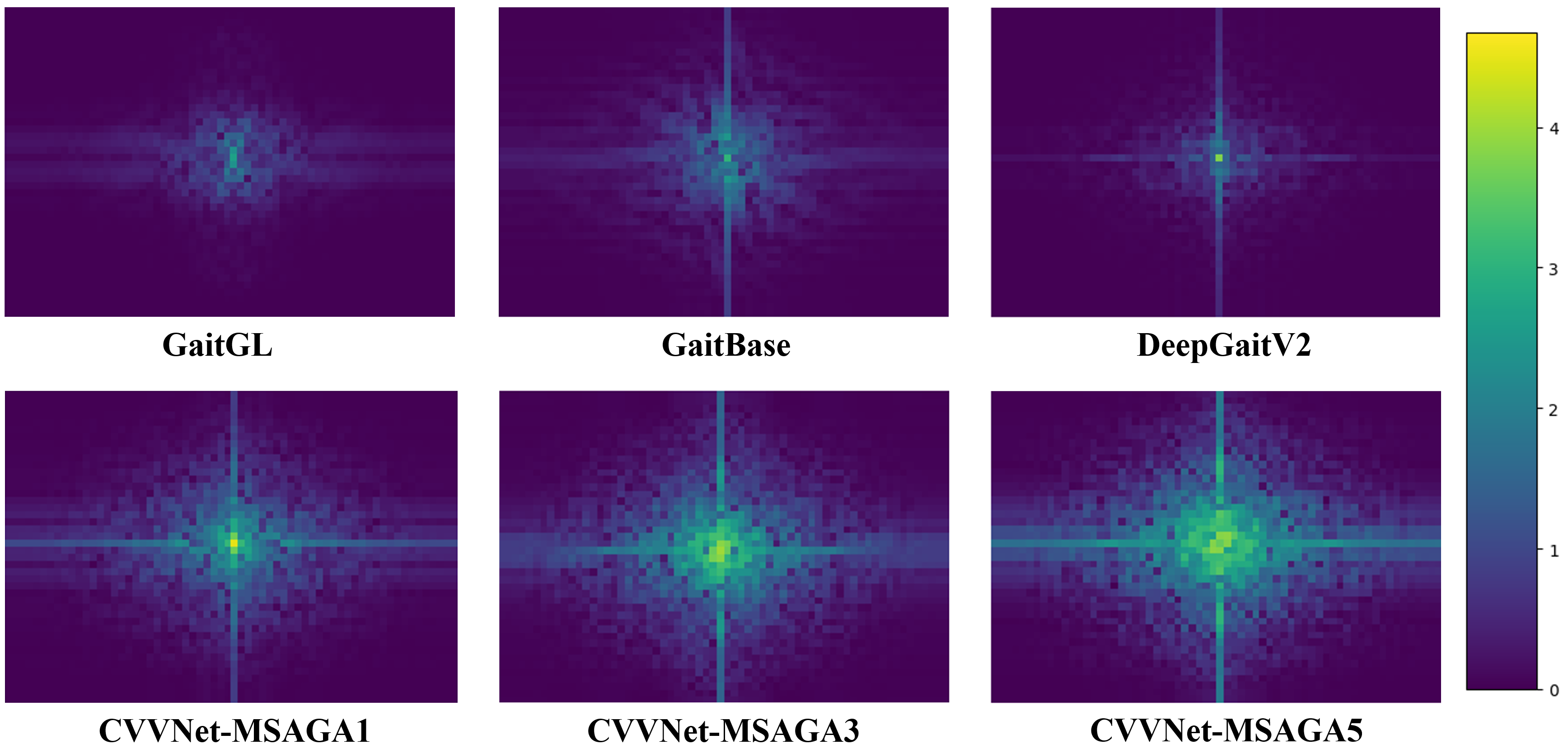}
\end{center}
   \caption{Frequency Spectrum Analysis of Different Gait Feature Extraction Methods. CVVNet-MSAGA1,3,5 indicates the module is in the 1st, 3rd, and 5th layers of Stage 2 in the model.}
   \vspace{-3mm}
\label{fig:Visualization}
\end{figure}

We draw the heatmap to highlight the viewpoint-specific feature activation in Fig.~\ref{fig:heatmap}. CVVNet dynamically prioritizes lower-body regions at low vertical angles and shifts attention to upper-body zones at high angles. Specifically, in low vertical views, the network focuses on dynamic details of the legs and feet, capturing crucial gait patterns that distinguish individual walking styles. In contrast, under high vertical views, the attention shifts to the head and torso, where distinctive features are more pronounced. This ability to adaptively adjust feature extraction priorities ensures that CVVNet effectively integrates multi-scale features, enhancing its robustness across varying perspectives and its generalization in complex environments. The spectral analysis is shown in Fig.~\ref{fig:Visualization}, which reveals distinct frequency characteristics: conventional convolution-based methods (GaitGL, GaitBase, DeepGaitV2) exhibit limited frequency diversity with concentrated central energy. In contrast, our CVVNet's MSAGA architecture demonstrates superior multi-scale frequency extraction, capturing both low- and high-frequency components. Increasing MSAGA layers progressively enriches the spectral content, enabling more comprehensive gait representations.


\section{Conclusion}

In this paper, we designed CVVNet, a novel cross-vertical-view network for addressing gait recognition across different vertical perspectives. CVVNet is composed of Multi-Scale Attention Gated Aggregation (MSAGA) modules and P3D blocks, adaptively extracts multi-scale gait frequency representations. It effectively separates and extracts low-frequency global structural information and high-frequency local dynamic details from gait silhouettes, fusing these features via a context-aware gating mechanism. Experiments on the DroneGait and Gait3D datasets validated its effectiveness, showing significant potential for practical applications. Our future work will focus on enhancing CVVNet by tackling more real-world challenges and integrating advanced technologies to improve its performance and robustness.

\section{Acknowledgments}This work was supported in part by the National
Natural Science Foundation of China (No. 62371142, Grant No. 62273107) and Guangdong Basic and Applied Basic Research Foundation (No. 2024A1515010404). 

\bibliographystyle{IEEEtran}

\end{document}